\documentclass[11pt,twocolumn]{article}
\usepackage{times}
\usepackage{graphicx}
\usepackage{amsmath}
\usepackage{amssymb}

\def \imagecenter#1{$\vcenter{\null\hbox{#1}}$}

\date{}

\usepackage{floatrow} 
\usepackage{subfig}
\usepackage{capt-of}

\usepackage{hyperref}

\begin{document}

\title{Multi-View Image-to-Image Translation Supervised by 3D Pose}
\author{Idit Diamant*, Oranit Dror*, Hai Victor Habi, Arnon Netzer\\
Sony Semiconductor Israel\\
{\tt\small \{idit.diamant,oranit.dror,hai.habi,arnon.netzer\}@sony.com}\\
\small{*equal contribution}\\
}

\maketitle


\begin{abstract}
We address the task of multi-view image-to-image translation for person image generation. 
The goal is to synthesize photo-realistic multi-view images with  pose-consistency across all views.
Our proposed end-to-end framework is based on a joint learning of multiple unpaired image-to-image translation models, one per camera viewpoint. The joint learning is imposed by constraints on the shared 3D human pose in order to encourage the 2D pose projections in all views to be consistent. Experimental results on the CMU-Panoptic dataset demonstrate the effectiveness of the suggested framework in generating photo-realistic images of persons with new poses that are more consistent across all views in comparison to a standard Image-to-Image baseline.
The code is available at: 
\url{https://github.com/sony-si/MultiView-Img2Img}.
\end{abstract}


\section{Introduction}

A wide range of approaches have been suggested for person image generation. Most methods employ deep generative models, such as Variational Auto-Encoders (VAE) \cite{kingma2014autoencoding} and Generative Adversarial Networks (GANs) \cite{NIPS2014_5423} due to their great performance in photo-realistic image synthesis. For example, VAE is a key component in \cite{lassner2017generative} for generating entirely new people with realistic clothing. Another representative work is \textit{VariGAN}, which is a combination of both VAEs and GANs for synthesizing realistic-looking images of a person captured at different views from a single input image \cite{zhao2018multiview}. This research field is of great value for a variety of applications, ranging from movie production to data augmentation,  alleviating data insufficiency in tasks, such as pose-estimation and person re-identification (Re-ID) \cite{zheng2017unlabeled,qian2018posenormalized}. 

The specific task of full-body image generation of persons is exceptionally challenging since human bodies are non-rigid objects with many degrees of freedom, resulting in a variety of possible deformation and articulation. 
Therefore, most approaches to tackle this task employ pose guidance. For example, similarly to \textit{VariGAN} \cite{zhao2018multiview} mentioned above, Si \textit{et al.} in \cite{Si2018pose} aim to synthesize novel view images of a person from a single input image. However, in contrast to \textit{VariGAN}, they manage to keep the human posture unchanged due to the utilized pose information. Typically, the pose information is employed in the form of anatomical keypoint positions, such as eyes, shoulders and hips \cite{ma2018pose,Si2018pose,balakrishnan2018synthesizing,Siarohin2018deformable,Bem_2018_ECCV_Workshops,Rodrigo_2019,chan2019everybody}. For instance, keypoint-based pose stick figures are used in \cite{chan2019everybody} as an intermediate representation to transfer full-body poses in videos from a source to a target person.

Most pose-guided methods are based on image-to-image translation networks that learn the mapping between a source and a target image. Usually the training is performed in a fully-supervised manner, based on samples of pose-corresponding image pairs \cite{ma2018pose, siarohin2019appearance, tang2020xinggan, huang2020generating, chan2019everybody}. However, for many tasks such a paired training dataset is unavailable or limited. Therefore, several pose-guided approaches try to tackle this task in an unsupervised manner \cite{pumarola2018unsupervised,chen2019unpaired,song2019unsupervised}. These approaches overcome the need for pose-corresponding image pairs by incorporating cycle-consistency \cite{CycleGAN2017} in their training objective.

Here, we address the task of image-to-image translation of a foreground person in a multi-camera scenario with pose-consistency across all views. In addition to the aforementioned applications for the single-view case, this task may have extensive practical applications especially as a mean of data-augmentation in cross-view problems (\textit{e.g.} cross-view person Re-ID), 3D action recognition and 3D human pose estimation \cite{chen2017synthesizing}. 
Compared to the single-view case, the task possesses an additional challenge of view-coherent. In particular, the goal is to synthesize photo-realistic images, one per camera viewpoint, of the target person in new poses that are consistent across all views. This means that the 2D human poses in the fake images are non-conflicting projections of the new 3D pose of the person. 

We tackle this task by employing 3D pose supervision for preserving view-consistency. The approach is based on unpaired image-to-image translation, which aims to learn the mapping between different but related visual domains in the absence of corresponding samples \cite{CycleGAN2017,NvidiaGAN2017,XGAN_2018}. In the specific task of interest, we have a pair of persons captured independently by the same set of synchronized and calibrated cameras. Each person thus has a multi-view image domain with tuples of images captured at the same time by all cameras. The two domains possess an underlying relationship of filming a similar scene of a foreground person in a variety of human-body poses, but with different appearance details (\textit{e.g.} clothing, hair-style and body shape). 

By learning this underlying relationship, the cross-mapping between the two domains can generate fake images of the persons in new poses. Specifically, translating a multi-view tuple of images from a source to a target domain will replace its domain-specific features (person appearance) with the ones of the target domain but preserve the domain-invariant features (human pose). This new multi-view tuple of images will capture the target person in the spatial pose of the source person. 

As stated above, a great challenge is to ensure that the 2D human poses depicted in the fake multi-view tuple of images are consistent projections of the new spatial pose of the target person. To withstand this challenge, we suggest a framework in which multiple unpaired image-to-image translation models (one per view) \cite{CycleGAN2017} are learned in a joint manner by imposing 3D constraints on the shared human pose.  

\noindent
\paragraph{} 
To summarize, the contributions of this paper are:
\begin{itemize}
    \item We propose a novel end-to-end solution for tackling the problem of multi-view pose-based image generation. The method is based on a joint learning of multiple unpaired image-to-image translation models (one per view). The joint-learning is supervised by 3D constraints on the human pose to encourage pose-consistency across all views.
    \item We apply our method on the CMU-Panoptic dataset \cite{joo2016panoptic} and demonstrate its performance in comparison to an Image-to-Image baseline. The results show that the proposed method is effective in both capturing the visual appearance and the global 3D pose, which is ensured to be consistent in a multi-view manner.
\end{itemize}


\section{Related Work}

Here, we briefly review existing literature that is closely related to our proposed method.

\paragraph{Image-to-Image Translation.}

An image-to-image translation problem aims to learn a mapping from an input image to a target image. A general-purpose solution to automatically learn image-to-image translation was first suggested in a framework named \textit{pix2pix} \cite{isola2018imagetoimage}. The framework is based on Conditional GANs \cite{yoo2016pixellevel} and has achieved reasonable results for a wide variety of problems, including photo-to-map, sketch-to-image and night-to-day. Nevertheless, for many tasks, \textit{pix2pix} is limited due to the requirement to perform the training in a supervised manner by supplying samples of input-output image pairs, which might not always be available. 

To overcome the need for paired training data, unsupervised methods have been emerged \cite{CycleGAN2017,NvidiaGAN2017,XGAN_2018,DRIT, choi2020stargan}. These methods for unpaired image-to-image translation aim to learn the cross-mapping between two different but related image domains (\textit{e.g.} different appearances of the same underlying scene) in the absence of corresponding samples. The input for the unpaired task is a sampled set of images for each domain and the adversary training is performed at the set level, meaning that the optimal generator should synthesize fake images that are indistinguishable from real images of the target domain. 
Additionally, transitivity between the two domains is exploited as a cycle consistency loss to regularize the cross-mapping between them \cite{CycleGAN2017}.

\paragraph{Pose-guided Person Image Generation.}

The pose-guided person image generation task aims at generating photo-realistic images of a person given an input image of that person and a target pose. This is basically an image-to-image translation task that converts an input image of a person into a new image of that person with the desired pose. 
The task was first introduced by Ma \textit{et al.} \cite{ma2018pose} and attracted great attention, leading to other approaches for tackling this task \cite{Si2018pose, esser2018variational, siarohin2019appearance, Zhu2019Pose, tang2020xinggan, tang2020cycle, huang2020generating}. 
Ma \textit{et al.} \cite{ma2018pose} suggested a novel two-stage coarse-to-fine approach ($PG^2$) based on conditional GANs to generate person images in different poses. The approach is based on synthesizing a new image by conditioning it on a reference input image and a desired pose in the format of anatomical body keypoints. To improve the control on the generation process, the authors further extended their original work in \cite{ma2018disentangled} to disentangle the input into foreground, background and pose embedding features. 

Similarly to \cite{ma2018disentangled}, recent works also suggest to decouple the complex entanglements of person image factors with the aim to generate person images in a controllable way. For instance, in \cite{huang2020generating} an end-to-end framework to decouple and re-couple entanglement factors is presented. The framework first decouples the entanglements between appearance, pose, foreground, background, local details and global structures. These entanglement factors are then progressively re-coupled by a generator named Appearance-aware Pose Stylizer (APS). Another example is \textit{XingGAN} \cite{tang2020xinggan}, which is a GAN-based network with two generation branches for modeling both person appearance and shape information. The two branches mutually improve each other and together progressively synthesize images from both person shape and appearance embeddings.

\begin{figure*}[t]
\centering{\includegraphics[scale=0.55, trim={0cm 0 0 0}]{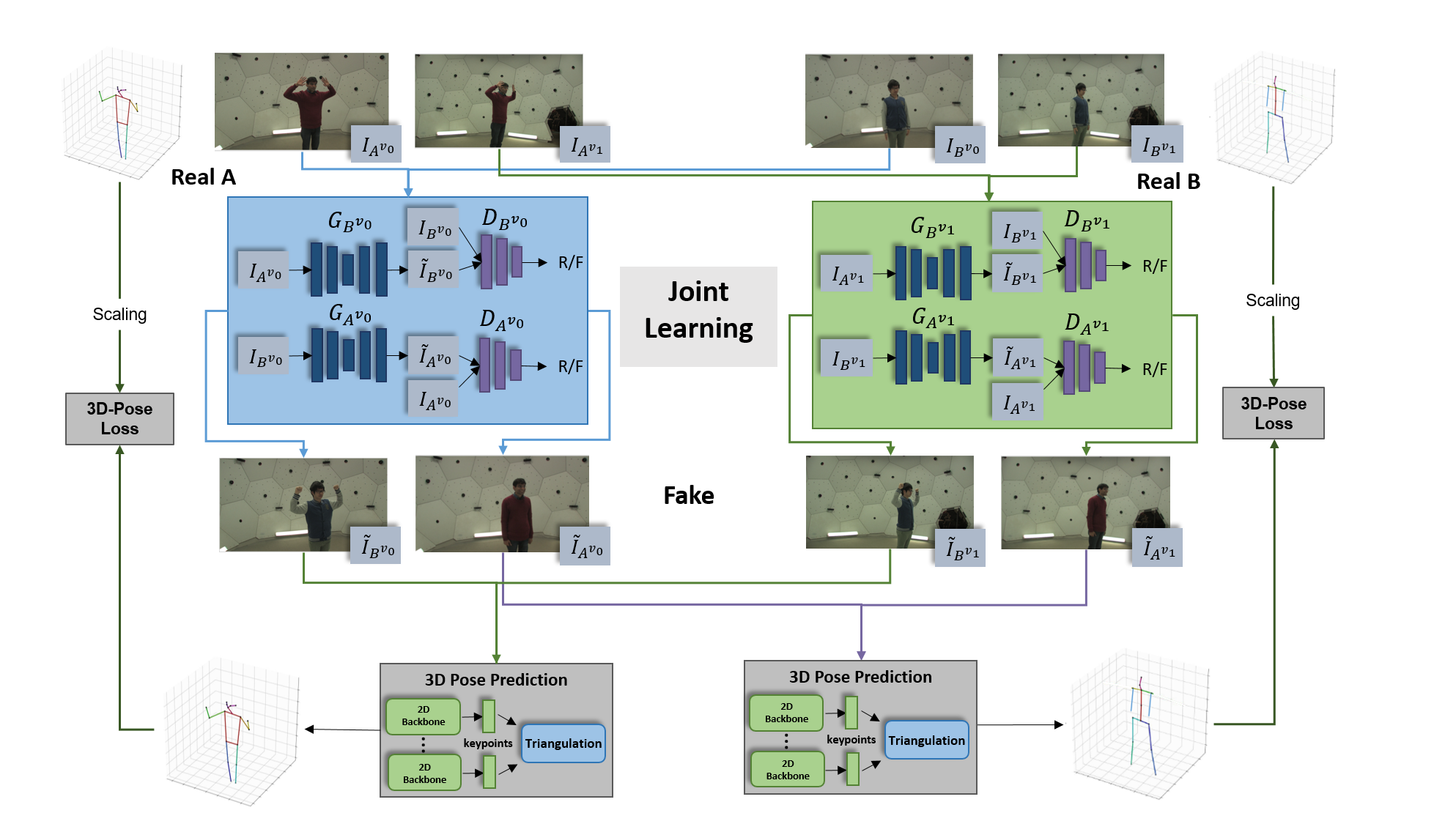}}
\caption{
\textbf{The overall network architecture for the dual-view scenario.} For each camera viewpoint, the architecture contains an unpaired image-to-image translation model \cite{CycleGAN2017} for learning the image mapping between a pair of persons. The two models are learned in a joint manner by shared constraints on the 3D human skeletons. These pose-guidelines encourage the fake images of each person to be view-consistent. } 
\label{Fig:architecture}
\end{figure*}

\paragraph{Unpaired Pose-guided Image Generation.}

The aforementioned pose-guided approaches learn their translation in a supervised manner by samples of pose-corresponding image pairs \cite{ma2018pose, siarohin2019appearance, tang2020xinggan, huang2020generating}. 
However, for many pose-guided tasks, a paired training dataset is unavailable or difficult to acquire.
Hence, several unsupervised approaches try to tackle the task in an unpaired manner \cite{pumarola2018unsupervised, chen2019unpaired, song2019unsupervised}. These approaches  avoid the need for corresponding pairs by incorporating cycle-consistency \cite{CycleGAN2017} in their objective. For instance, Pumarola \textit{et al.} in \cite{pumarola2018unsupervised} combine concepts adopted from CycleGAN \cite{CycleGAN2017} and style-transfer \cite{Gatys_2016_CVPR}, whereas \textit{UPG-GAN} \cite{chen2019unpaired} combines elements from both CycleGAN \cite{CycleGAN2017} and VAE \cite{esser2018variational}. 

\paragraph{Pose Estimation.}
Most methods for 3D human pose estimation are two-tier. They first predict 2D skeletons and in an additional stage they estimate the 3D pose from these 2D skeletons \cite{iskakov2019learnable}.
The state-of-the-art methods are based on multi-view data \cite{iskakov2019learnable, tu2020voxelpose, chen2020multiperson}. 
In this paper, we use the network of \cite{iskakov2019learnable} for multi-view 3D pose estimation. This method is based on learnable triangulation, which allows to reduce the number of views required for accurate prediction of 3D pose.

\section{Method}

We assume a set of cameras with known projection matrices that capture a foreground person simultaneously. Given two persons, each independently filmed by the set of cameras, our method aims to find the cross-mapping between their image domains. The image domain of each person is multi-view, that is, containing tuples of images that have been captured simultaneously by all cameras. Once learned, the mapping can be used to generate fake images of the persons in a variety of new poses. Specifically, translating a multi-view tuple of images from a source to a target person will replace the person  appearance (domain-specific features) with the ones of the target person, while preserving the source human pose (domain-invariant features). This new multi-view tuple of images will capture the target person in the spatial pose of the source person. 

We suggest a framework in which multiple unpaired image-to-image translation models \cite{CycleGAN2017} (one per viewpoint) are learned in a joint manner by imposing spatial constraints on the shared 3D human pose. These constraints guide the joint-learning and encourage all translation models to generate for each person fake images that are view consistent, \textit{i.e.} their 2D human poses are consistent projections of the new target 3D pose. See Figure \ref{Fig:architecture} for an illustration of the dual-view case.

\paragraph{Problem Formulation.}

For simplicity, below we will formulate the problem for the dual-view case of a pair of cameras capturing a foreground person simultaneously.

For two given persons, $A$ and $B$, and two synchronized camera viewpoints, $v_0$ and $v_1$, the input contains for each person a set of dual-view pairs of images captured simultaneously by the two camera viewpoints. The goal is to learn the cross-mapping between the dual-view image domains of the two persons, namely $A^{v_0} \times A^{v_1} \leftrightarrow B^{v_0} \times B^{v_1}$, where $A^{v_0}$, $A^{v_1}$, $B^{v_0}$ and $B^{v_1}$ are the respective single-view image domains. Once learned, the mapping of a dual-view pair of images from a source to a target domain will replace its domain-specific features with the ones of the target domain, but preserves the domain-invariant features. Therefore, when translating for example $(I_{A^{v_0}}, I_{A^{v_1}}) \in A^{v_0} \times A^{v_1}$ to $(I_{B^{v_0}}, I_{B^{v_1}}) \in B^{v_0} \times B^{v_1}$, the pose of the target person $B$ in the fake pair of images $(I_{B^{v_0}}, I_{B^{v_1}})$ will be spatially consistent and as similar as possible to the pose of person $A$ in the source real images, $(I_{A^{v_0}}, I_{A^{v_1}})$, up to the difference in body proportions between the two persons.

To ensure that the dual-view mapping will preserve pose consistency, 3D information on the human pose is taken into account during training. Therefore, each dual-view pair of images in the input is provided with the respective 3D pose of the filmed person. The 3D human pose is represented as a spatial composition of a fixed set of anatomical body keypoints with indices $j \in (1,...,J)$, that is, $P=\langle p_1,...,p_J \rangle$, where $p_j=(x_j, y_j, z_j) \in \mathbb{R}^3$ is the 3D coordinates of joint $j$. Overall, the input for person $A$ is a collection of $(I^t_{A^{v_0}}, I^t_{A^{v_1}}, P^t_A) \in A^{v_0} \times A^{v_1} \times P_A$, where $P_A$ is the pose domain of person $A$ and $t$ is a frame index. A similar input is provided for person $B$, that is a collection of $(I^t_{B^{v_0}}, I^t_{B^{v_1}}, P^t_B) \in B^{v_0} \times B^{v_1} \times P_B$.

\paragraph{Framework.}
The framework consists of two symmetric modules for unpaired image-to-image translation \cite{CycleGAN2017}, one per camera viewpoint. Each module learns the cross-domain mapping between images captured by the respective camera viewpoint of person $A$ to person $B$ and vice versa, that is, $A^{v_0} \leftrightarrow B^{v_0}$ and $A^{v_1} \leftrightarrow B^{v_1}$. During training, a joint learning of the two translation modules is supervised by shared 3D constraints on the human poses. The aim is to encourage synthesis of fake images that are pose-consistent across all views. The shared spatial constraints are imposed between the ground-truth 3D skeleton of the source person and a predicted 3D skeleton of the target person. The overall network architecture is presented in Figure \ref{Fig:architecture}. 

\paragraph{Single-View Image-to-Image Translation.}
For each camera viewpoint $v \in \{v_0,v_1\}$, the architecture contains an unpaired image-to-image translation module. The goal is to learn the cross-domain mapping between two unpaired image domains $A^v \leftrightarrow B^v$. This mapping translates a single image of a person to a fake image of the other person by replacing the person appearance, but preserving the pose of the source person. 
The implementation is based on CycleGAN \cite{CycleGAN2017}. As such, the module consists of two image generators, $G_{A^v}: B^v \to A^v $ and $G_{B^v}: A^v \to B^v $, as well as two adversarial discriminators, $D_{A^v}$ and $D_{B^v}$. The generators are the mapping functions between the domains while the discriminators distinguish between real and fake images for the target domain. A \textit{generative adversarial loss} \cite{NIPS2014_5423} is used to encourage the mappings to produce realistic fake images that are visually similar to real images from the target domain. For the $A^v \to B^v $ mapping this loss is:

\begin{equation}
\begin{aligned}
& L^v_{gan}(G_{B^v}, D_{B^v}) = \mathbb{E}_{I_{B^v} \sim B^v}  \log{D_{B^v}(I_{B^v})} \\ 
& + \mathbb{E}_{I_{A^v} \sim A^v} \log{(1- D_{B^v}(G_{B^v}(I_{A^v})))} 
\end{aligned}
\label{eq:gan}
\end{equation}

The cross-domain translation is further regularized via a cycle-consistency loss:

\begin{equation}
\begin{aligned}
& L^v_{cycle}(G_{A^v}, G_{B^v}) = \\
& \mathbb{E}_{I_{A^v} \sim A^v} [||G_{A^v}(G_{B^v}(I_{A^v})) - I_{A^v} ||] + \\
& \mathbb{E}_{I_{B^v} \sim B^v} [|| G_{B^v}(G_{A^v}(I_{B^v})) - I_{B^v} ||] 
\end{aligned}
\label{eq:cycle-consistency}
\end{equation}

\paragraph{} Last, we use the \textit{identity loss} \cite{CycleGAN2017} to encourage each generator to be the identity mapping when real images of the target domain are provided as its input. Specifically, the identity loss for the $G_{A^v}$ generator is:
\begin{equation}
\begin{aligned}
L^v_{identity}(G_{A^v}) = \mathbb{E}_{I_{A^v} \sim A^v} [||G_{A^v}(I_{A^v}) - I_{A^v} ||] 
\end{aligned}
\label{eq:identity}
\end{equation}

\noindent
Overall, the objective for viewpoint $v$ is:
\begin{equation}
\begin{aligned}
& L^v(G_{A^v}, G_{B^v}, D_{A^v}, D_{B^v}) = \lambda_{1} L^v_{cycle}(G_{A^v}, G_{B^v}) + \\ 
& \lambda_{2} [L^v_{gan}(G_{A^v}, D_{A^v}) + L^v_{gan}(G_{B^v}, D_{B^v})] + \\ 
& \lambda_{3}[L^v_{identity}(G_{A^v})+L^v_{identity}(G_{B^v})] 
\end{aligned}
\label{eq:sinlge-view-loss}
\end{equation}

\noindent where $\lambda_{1}$, $\lambda_{2}$ and $\lambda_{3}$ are hyper-parameters that control the relative importance of the three objectives. 

\paragraph{Joint Learning with 3D Pose Constraints.} 

The two image-to-image translations (one per camera viewpoint) are learned jointly with a shared loss term in the full objective function. The full objective function consists of the loss term specified in Equation \ref{eq:sinlge-view-loss} for each view $v \in \{v_0, v_1\}$ as well as terms that impose 3D constraints on the human poses. Our aim is to ensure that the 2D poses of the target person in the two fake images are view-consistent and as similar as possible to the 2D poses of the source person in the real images. 

To this end, we predict the 3D pose of the target person based on the dual-view pair of fake images. Practically, this human pose is predicted by a pre-trained model of a state-of-art network for multi-view 3D pose estimation \cite{iskakov2019learnable}. We then compare the predicted 3D pose of the target person to a scaled version of the ground-truth 3D pose of the source person. The pose scaling is required to make sure that the comparison is invariant to body proportions. Last, the 3D distance between the two poses is incorporated as a joint-learning loss term in the full objective function (see Figure \ref{Fig:architecture}). In our implementation, the distance between the poses is computed as a smooth Mean Square Error (MSE) in order to be more robust to outliers at training time \cite{iskakov2019learnable}:

\footnotesize
\begin{equation}
\begin{aligned}
& MSE_{smooth}(P_i, P_j) = \left\{
                \begin{array}{ll}
                MSE(P_i,P_j), \; if \; MSE<\epsilon \\
                \vspace{0.1mm}\\
                MSE(P_i,P_j)^{0.1} \cdot \epsilon^{0.9}, \;o/w
                \vspace{0.1mm}\\
                \end{array}
                \right.
\end{aligned}
\label{eq:smooth_MSE}
\end{equation}
\normalsize

\paragraph{}
\noindent where $P_i$ and $P_j$ are poses and $\epsilon$ is a predefined threshold.

\paragraph{}The $3D$ loss term for the $A^{v_0} \times A^{v_1} \to B^{v_0} \times B^{v_1}$ mapping is:
\begin{equation}
\begin{aligned}
& L_{3D}(G_{B^{v_0}}, G_{B^{v_1}}) = MSE_{smooth}(S_{A \to B}(P^t_A),\\ 
& P_{3D}(G_{B^{v_0}}(I^t_{A^{v_0}}), G_{B^{v_1}}(I^t_{A^{v_1}})))
\end{aligned}
\label{eq:3d_loss}
\end{equation}

\noindent where $(I^t_{A^{v_0}}, I^t_{A^{v_1}})$ is a pair of real images in $A^{v_0} \times A^{v_1}$, $P^t_A$ is the associated ground-truth 3D pose, $S_{A \to B}$ is the pose-scaling function and $P_{3D}$ is a pre-trained multi-view 3D pose-estimation network model \cite{iskakov2019learnable}. 

\paragraph{}A $3D$ loss term for the $B^{v_0} \times B^{v_1} \to A^{v_0} \times A^{v_1}$ mapping is defined similarly and the full objective is therefore:
\begin{equation}
\begin{aligned}
& L_{total} = \sum_{i=0,1} L^{v_i}(G_{A^{v_i}}, G_{B^{v_i}}, D_{A^{v_i}}, D_{B^{v_i}}) \; + \\
 & \lambda_{4} [L_{3D}(G_{A^{v_0}}, G_{A^{v_1}}) + L_{3D}(G_{B^{v_0}}, G_{B^{v_1}})]
\end{aligned}
\label{eq:single_view_loss}
\end{equation}

\noindent where $\lambda_{4}$ controls the relative importance of the 3D loss compared to the other losses.
Also, as in all GANs, during training we aim to solve a min-max problem trying to find the optimal point of the generators and discriminators.

\section{Experimental Results}
We evaluate the proposed method through experiments conducted on the CMU-Panoptic dataset \cite{joo2016panoptic} and compare our results with CycleGAN \cite{CycleGAN2017}, a standard image-to-image translation baseline. 

\vspace{0.1cm}
\subsection{Implementation Details}
\vspace{0.1cm}

\paragraph{Network and Training Details.} We have adopted the architecture of Zhu \textit{et al.} for our generative networks \cite{CycleGAN2017}. For training, we use the Adam solver with a batch size of 1. The initial learning rate is 0.0002, which is used for the first 100 epochs. After 100 epochs we start a linear decay for 200 epochs till zero. Last, we apply translation augmentation in the x-axis only to avoid cropping the person's body.

\paragraph{The CycleGAN Baseline.} The baseline for comparison consists of a separated CycleGAN network  \cite{CycleGAN2017} for each camera viewpoint for translating images between a pair of humans, $A$ and $B$. We trained these two CycleGAN networks separately and compared their results to the ones achieved by the proposed network.

\vspace{0.2cm}
\subsection{Qualitative Results} 
\begin{figure}[t]
    \centering
    \begin{tabular}{ p{0.001\textwidth} p{0.9\textwidth} }
       \imagecenter{\footnotesize a} &  \imagecenter{\includegraphics[trim={0 -15 0 0}, clip, width=0.9\textwidth]{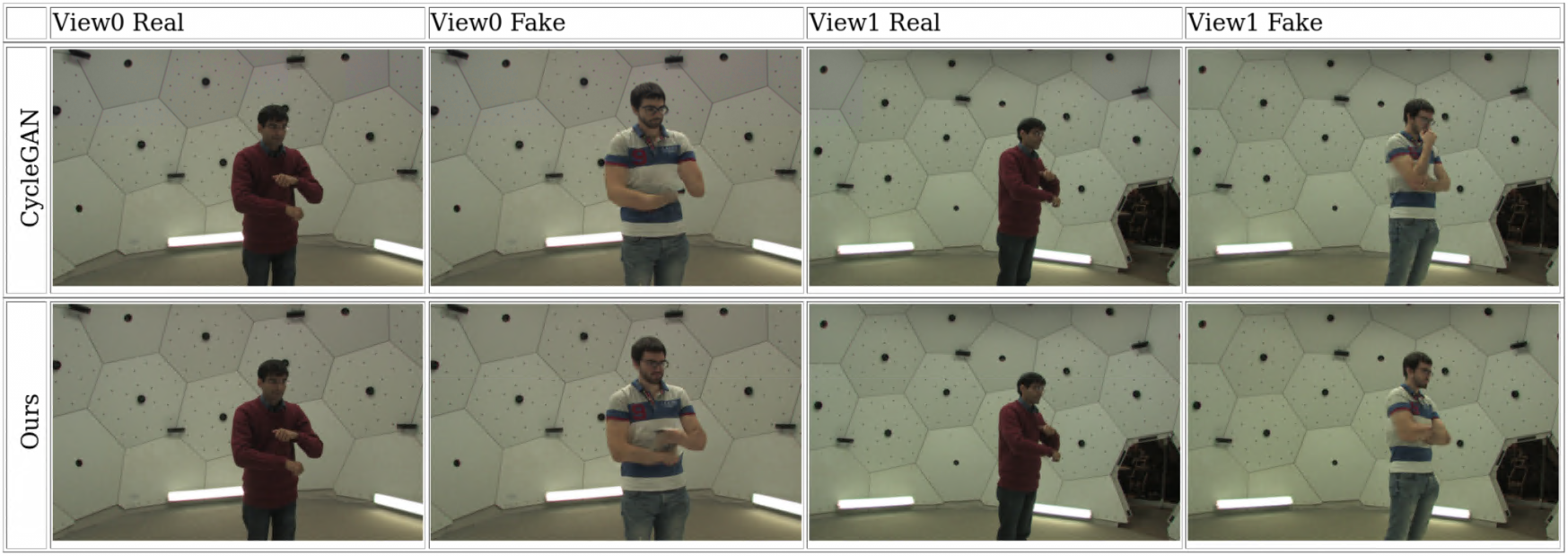}}\\
       \imagecenter{\footnotesize b} & \imagecenter{\includegraphics[trim={0 -15 0 30}, clip, width=0.9\textwidth]{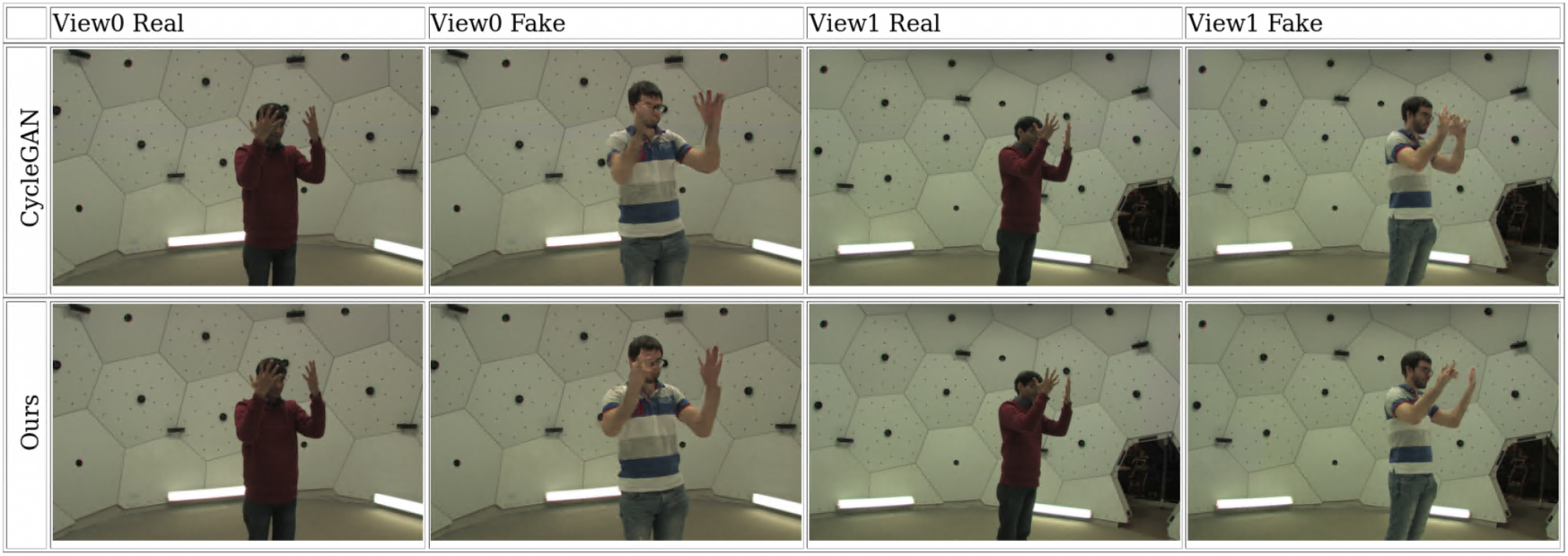}}\\
       \imagecenter{\footnotesize c} & \imagecenter{\includegraphics[trim={0 0 0 30}, clip, width=0.9\textwidth]{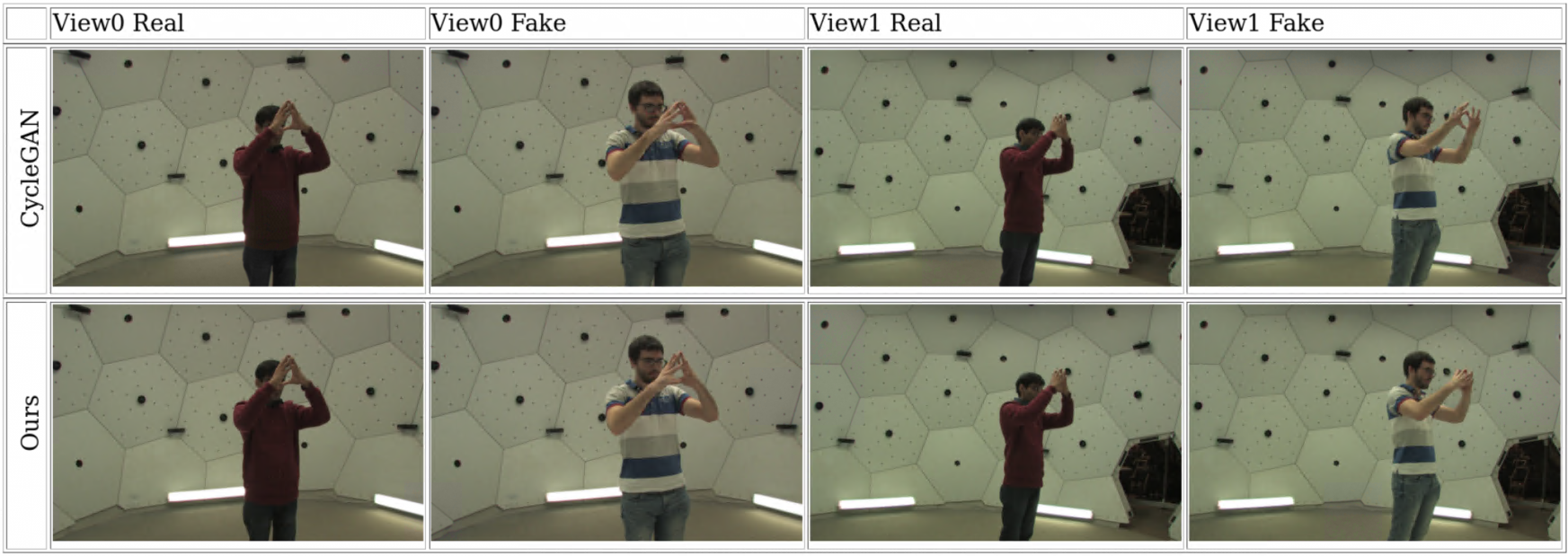}}\\
    \end{tabular} 
    \caption{\textbf{Pose consistency in person A$\to$B translation.} Results of our approach in comparison to CycleGAN \protect\cite{CycleGAN2017} demonstrate the effectiveness of the dual-view joint learning in preserving human pose.  
    } 
    \label{Fig:results_views_00_24_pair1_a2b_pose_preservation}
\end{figure}

\begin{figure}[t!]
    \centering
    \begin{tabular}{ p{0.001\textwidth} p{0.9\textwidth} }
        \imagecenter{\footnotesize a} &  \imagecenter{\includegraphics[trim={0 5 0 0}, clip, width=0.9\textwidth]{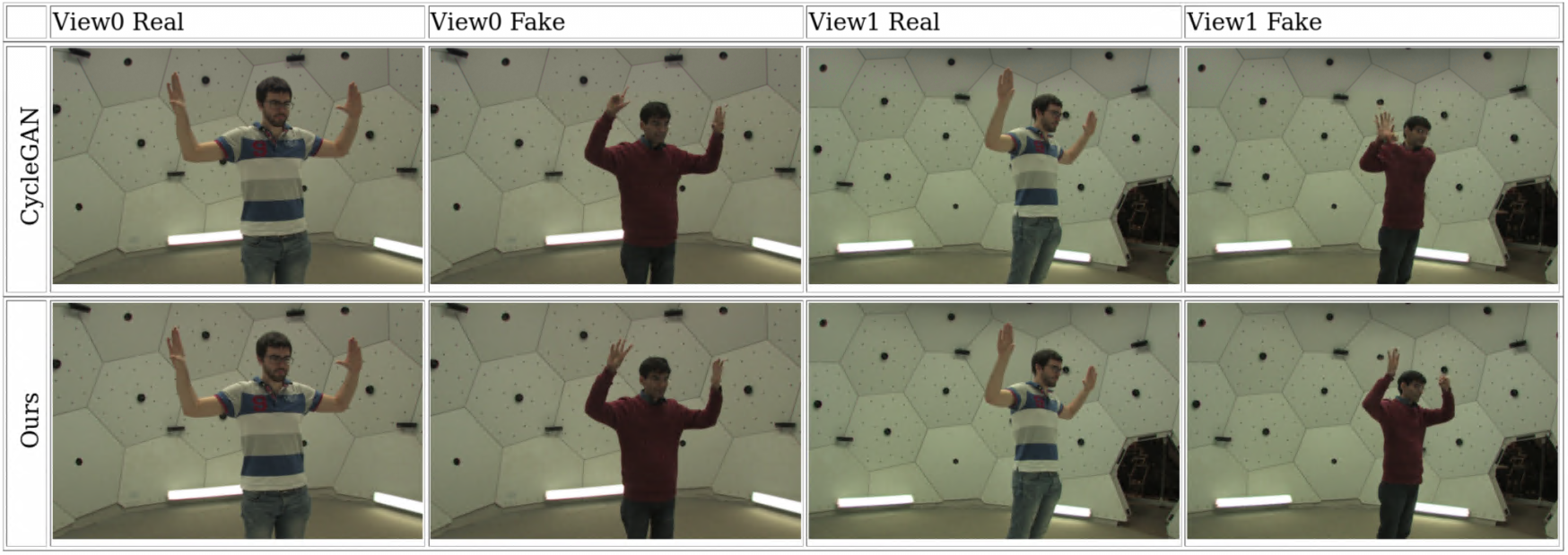}}
    \label{Fig:results_views_00_24_pair1_btoa_pose_preservation_a} \\
    \imagecenter{\footnotesize b} &  \imagecenter{\includegraphics[trim={0 -15 0 35}, clip, width=0.9\textwidth]{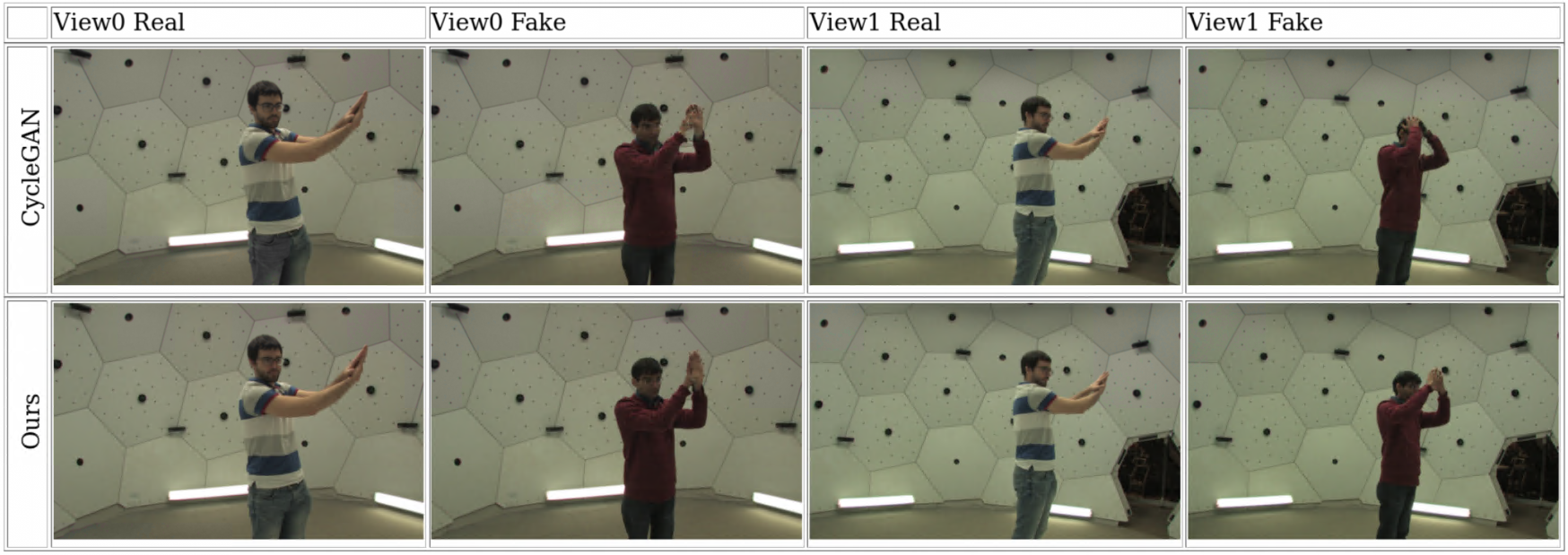} \label{Fig:results_views_00_24_pair1_btoa_pose_preservation_b}} \\
    \imagecenter{\footnotesize c} &  \imagecenter{\includegraphics[trim={0 0 0 30}, clip, width=0.9\textwidth]{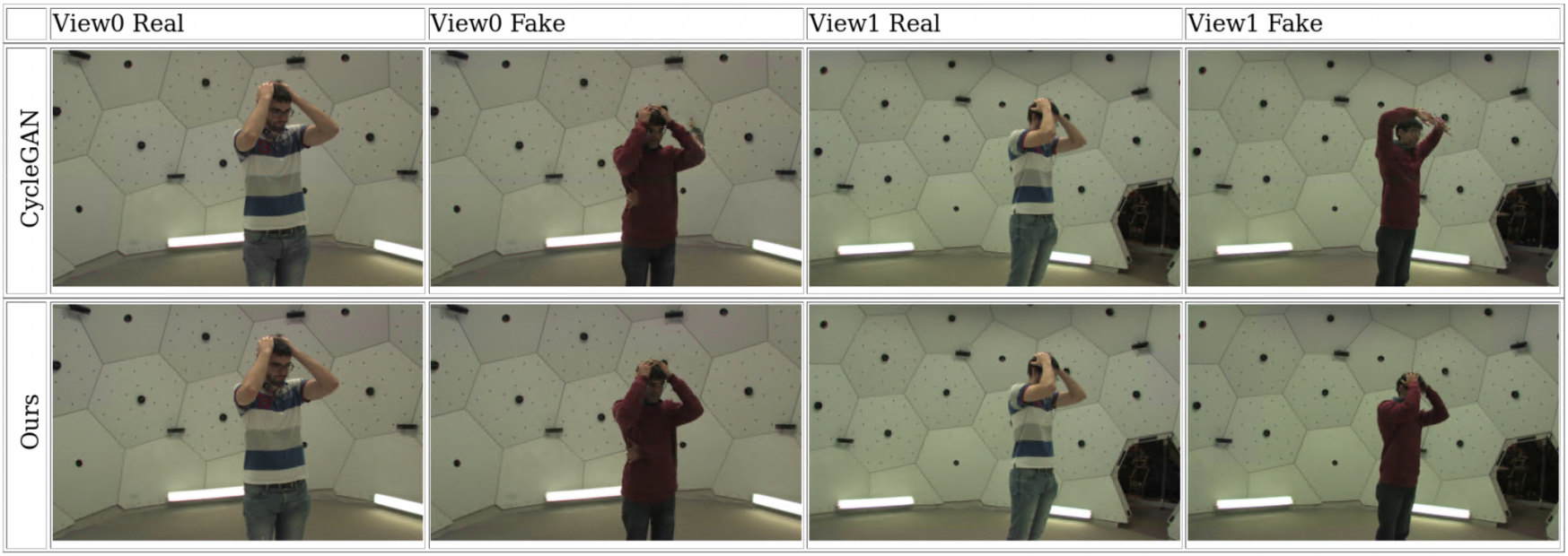} \label{Fig:results_views_00_24_pair1_btoa_pose_preservation_c}} \\
    \end{tabular}
\caption{\textbf{Pose consistency in person B$\to$A translation.} Results of our approach in comparison to CycleGAN \protect\cite{CycleGAN2017} demonstrate the effectiveness of the dual-view joint learning in preserving human pose. 
}
\label{Fig:results_views_00_24_pair1_btoa_pose_preservation}
\end{figure}

\begin{figure}[t]
   \centering
    \begin{tabular}{ p{0.001\textwidth} p{0.9\textwidth} }
        \imagecenter{\footnotesize a} &  \imagecenter{\includegraphics[trim={2 -15 0 0}, clip, width=0.9\textwidth]{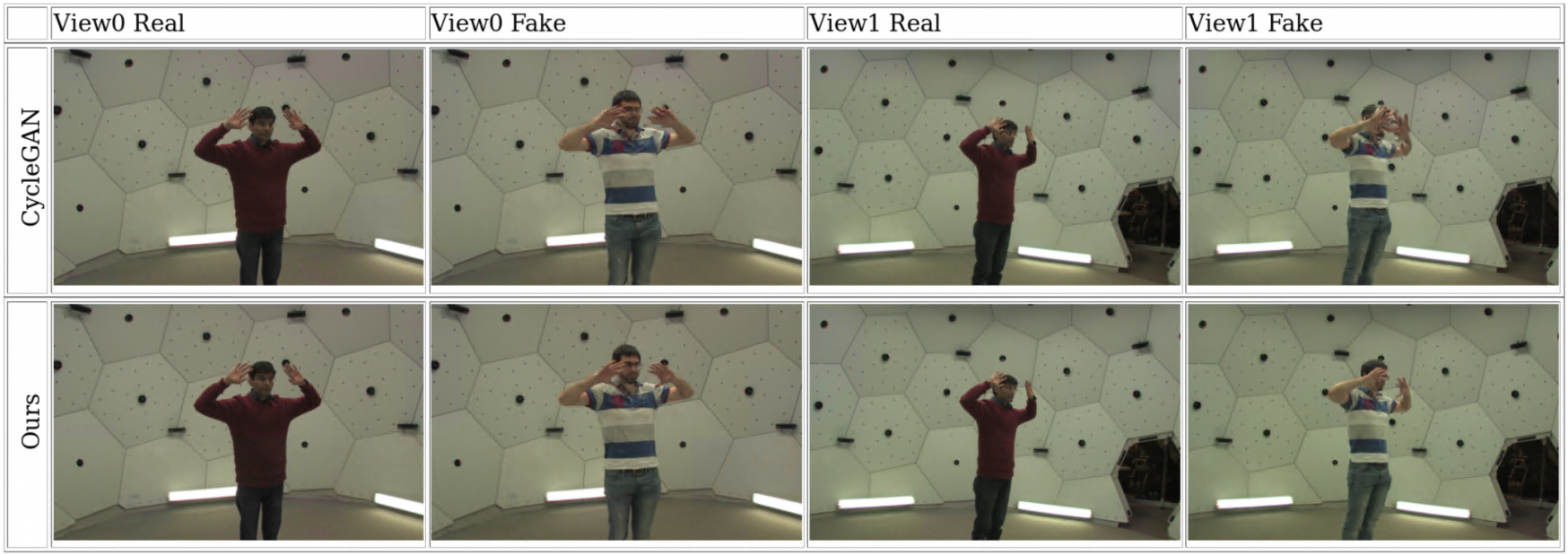}} \\
\imagecenter{\footnotesize b} &  \imagecenter{\includegraphics[trim={2 -15 0 40}, clip, width=0.9\textwidth]{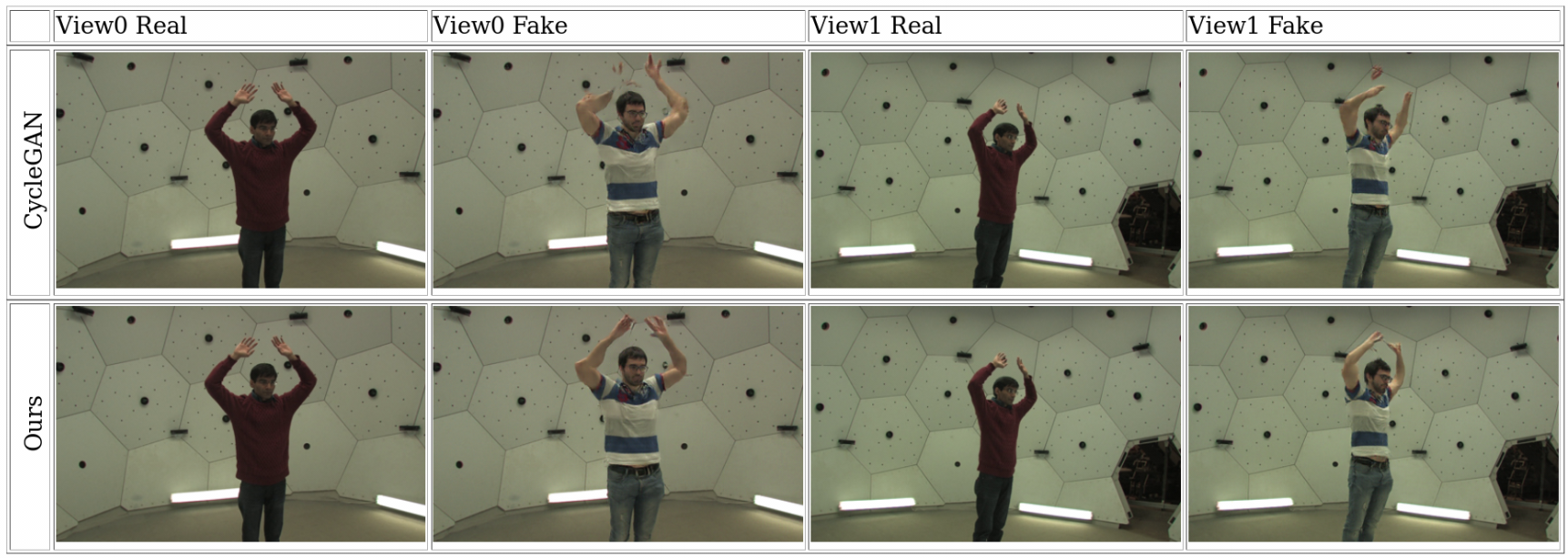}} \\
\imagecenter{\footnotesize c} &  \imagecenter{\includegraphics[trim={2 -15 0 40}, clip, width=0.9\textwidth]{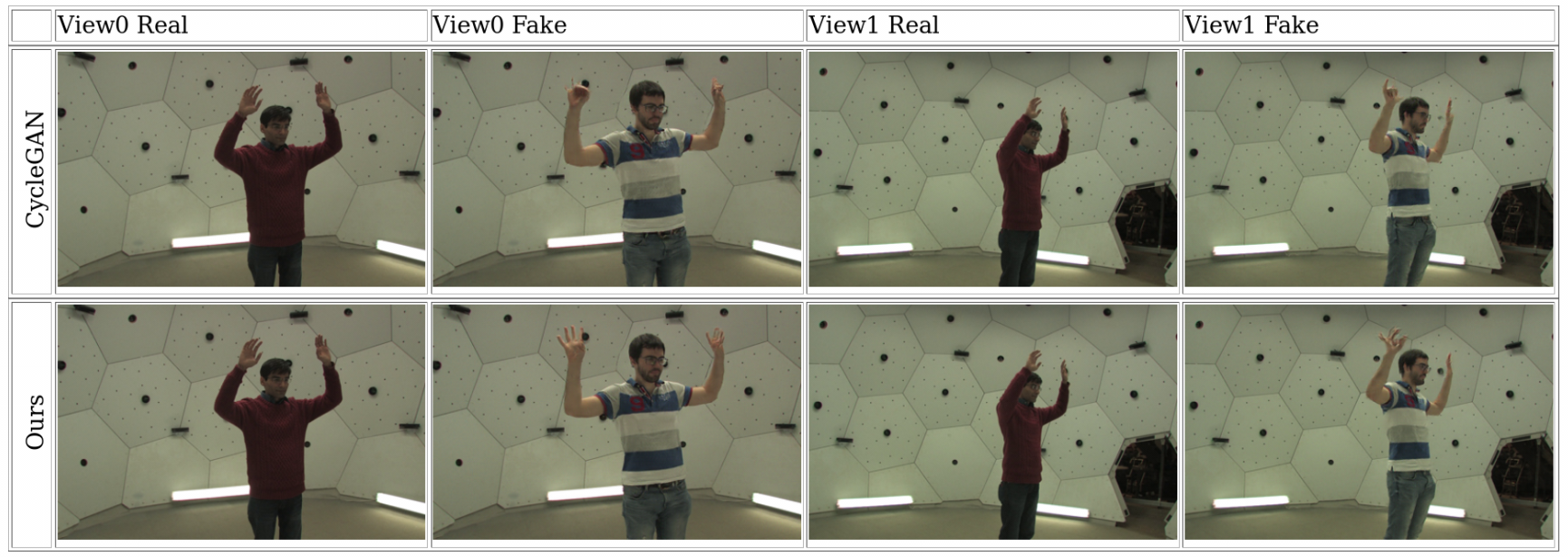}} \\
\imagecenter{\footnotesize d} &  \imagecenter{\includegraphics[trim={2 -15 0 40}, clip,  width=0.9\textwidth]{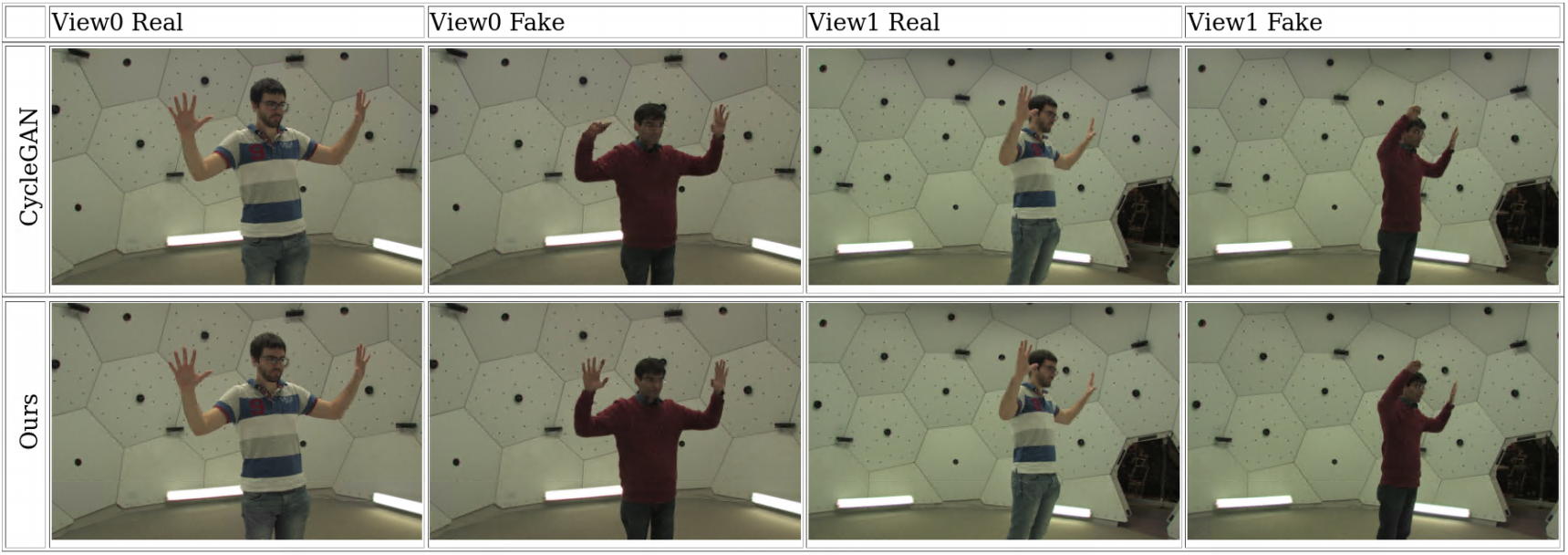}} \\
\end{tabular}
\caption{\textbf{Local feature preservation.} In all examples both approaches preserved the overall human pose. However, local features, such as fingers, are more consistent between real and fake images generated by our approach.} 
\label{Fig:results_views_00_24_pair1_a2b_feature_consistency}
\end{figure}

\begin{figure}[t]
\centering
    \begin{tabular}{ p{0.001\textwidth} p{0.9\textwidth} }
        \imagecenter{\footnotesize a} &  \imagecenter{\includegraphics[trim={0 -5 0 0}, clip, width=0.9\textwidth]{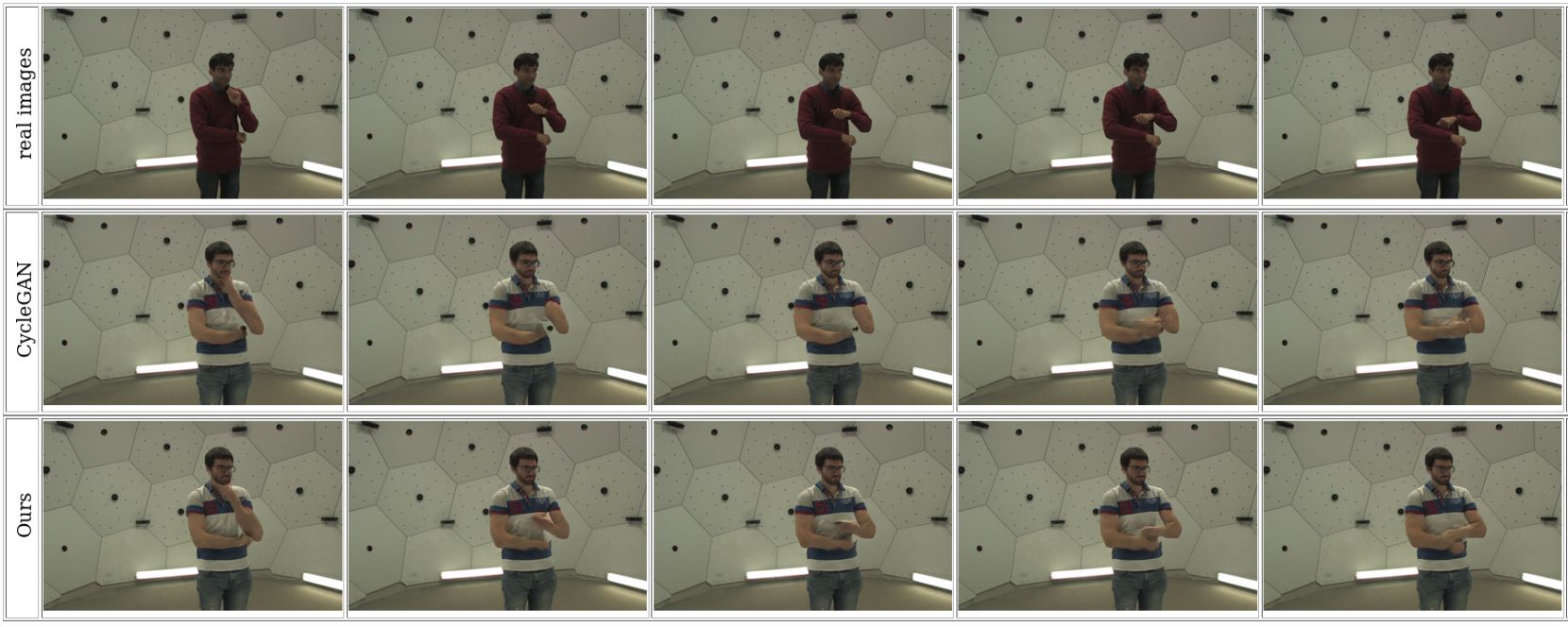} } \\
\imagecenter{\footnotesize b} &  \imagecenter{\includegraphics[trim={0 5 0 0}, clip, width=0.9\textwidth]{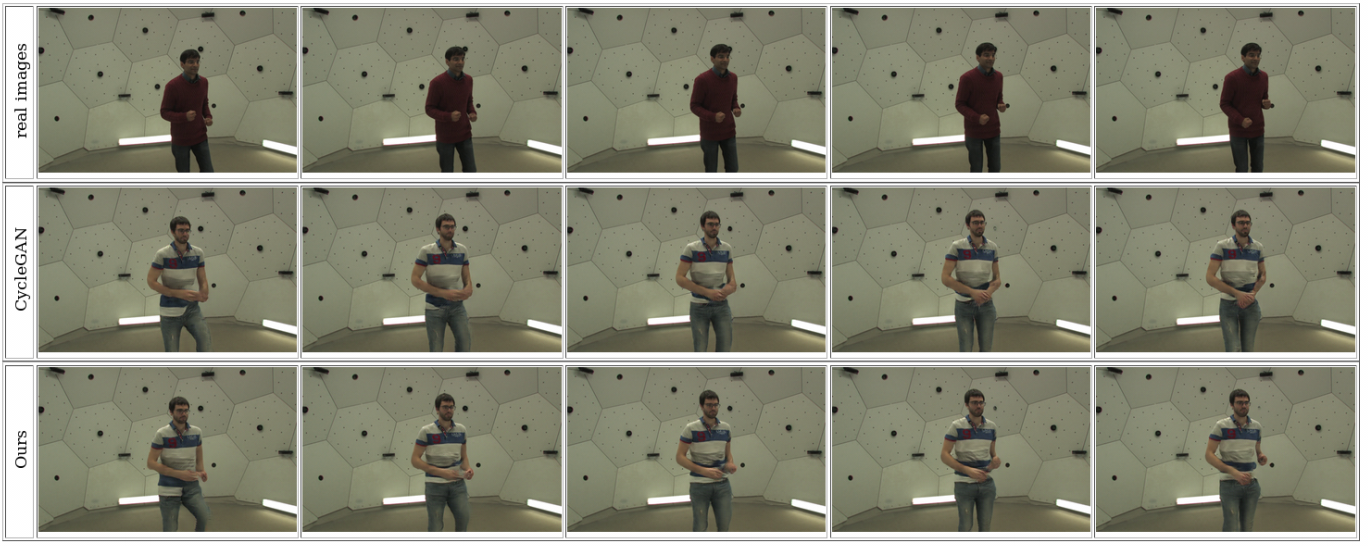} } \\
\imagecenter{\footnotesize c} &  \imagecenter{\includegraphics[trim={0 0 0 0}, clip, width=0.9\textwidth]{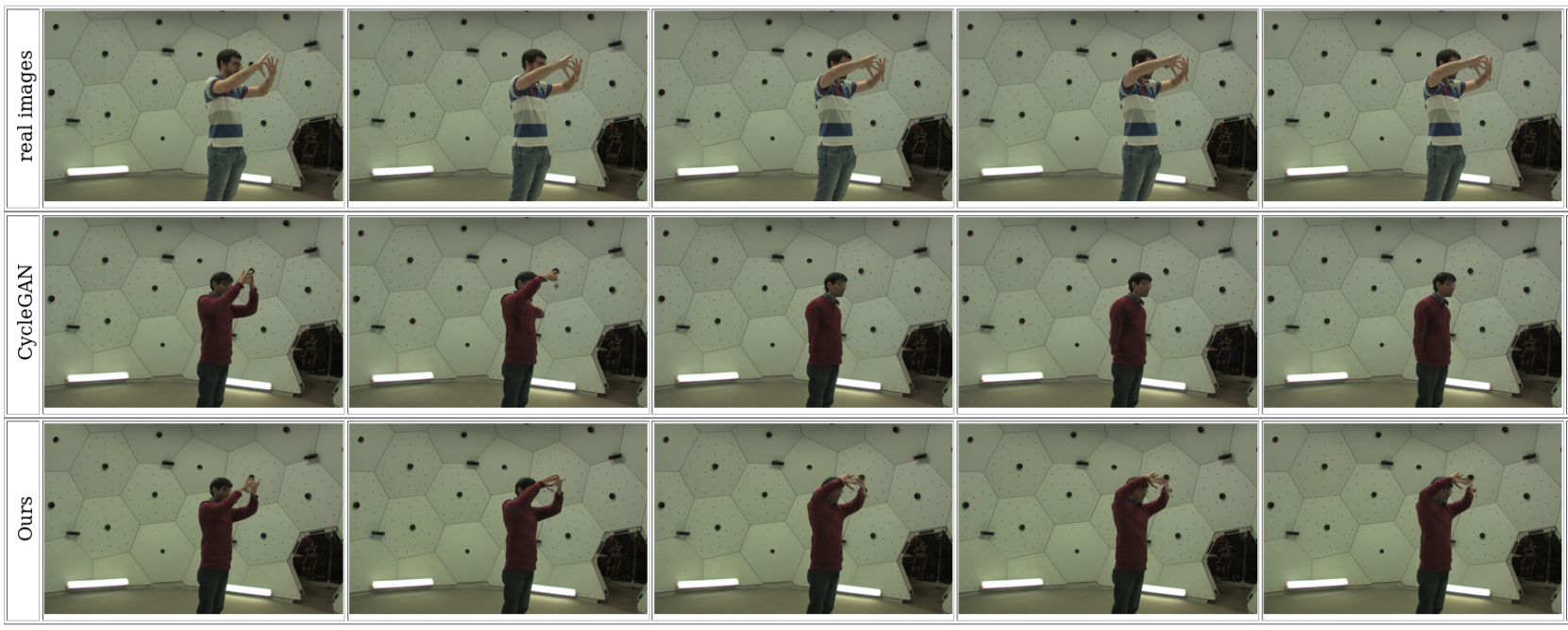} }
\end{tabular}
\caption{\textbf{Performance on video frame sequences}. Three sequences of real video frames are shown (a-c). For each one of them, a sequence of fake frames generated by our method is presented along with a sequence of fake frames generated by CycleGAN.}
\label{Fig:Sequence_comparison}
\end{figure}

\begin{figure}[t!]
\centering
    \begin{tabular}{ p{0.001\textwidth} p{0.9\textwidth} }
\imagecenter{\footnotesize a} &  \imagecenter{\includegraphics[trim={0 -2 0 0}, clip, width=0.9\textwidth]{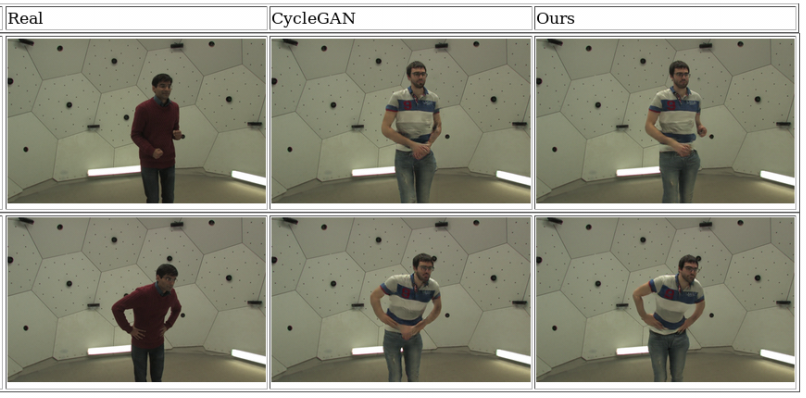}} \\
\imagecenter{\footnotesize b} &  \imagecenter{\includegraphics[trim={0 -2 0 0}, clip, width=0.9\textwidth]{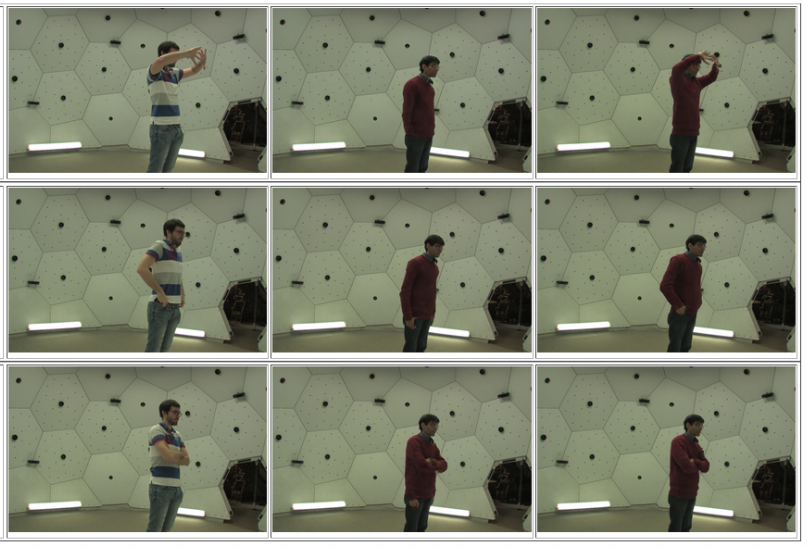}} \\
\imagecenter{\footnotesize c} &  \imagecenter{\includegraphics[trim={0 -2 0 0}, clip, width=0.9\textwidth]{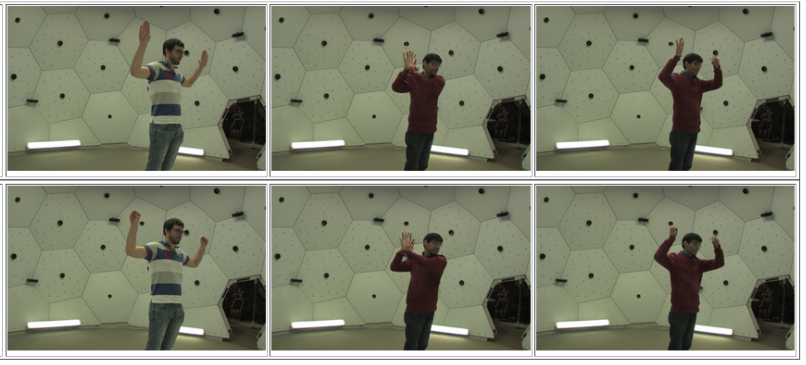}} \\
\end{tabular}
\caption{\textbf{Robustness and stability.} The examples show the robustness and stability of our approach in comparison to CycleGAN \protect\cite{CycleGAN2017}. } 
\label{Fig:results_CycleGAN_collapse_cases}
\end{figure}

\begin{figure}[t!]
\centering
    \begin{tabular}{ p{0.001\textwidth} p{0.9\textwidth} }
\imagecenter{\footnotesize a} &  \imagecenter{\includegraphics[trim={0 -15 0 0}, clip, width=0.9\textwidth]{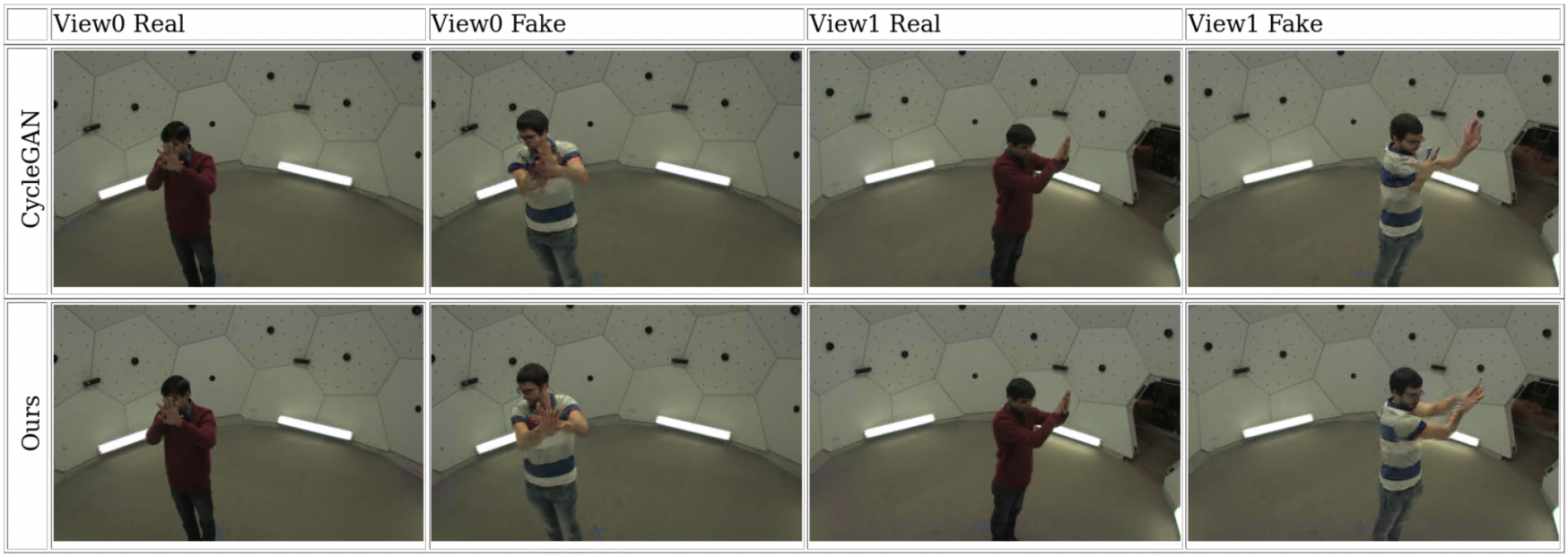}} \\
\imagecenter{\footnotesize b} &  \imagecenter{\includegraphics[trim={0 -15 0 40}, clip, width=0.9\textwidth]{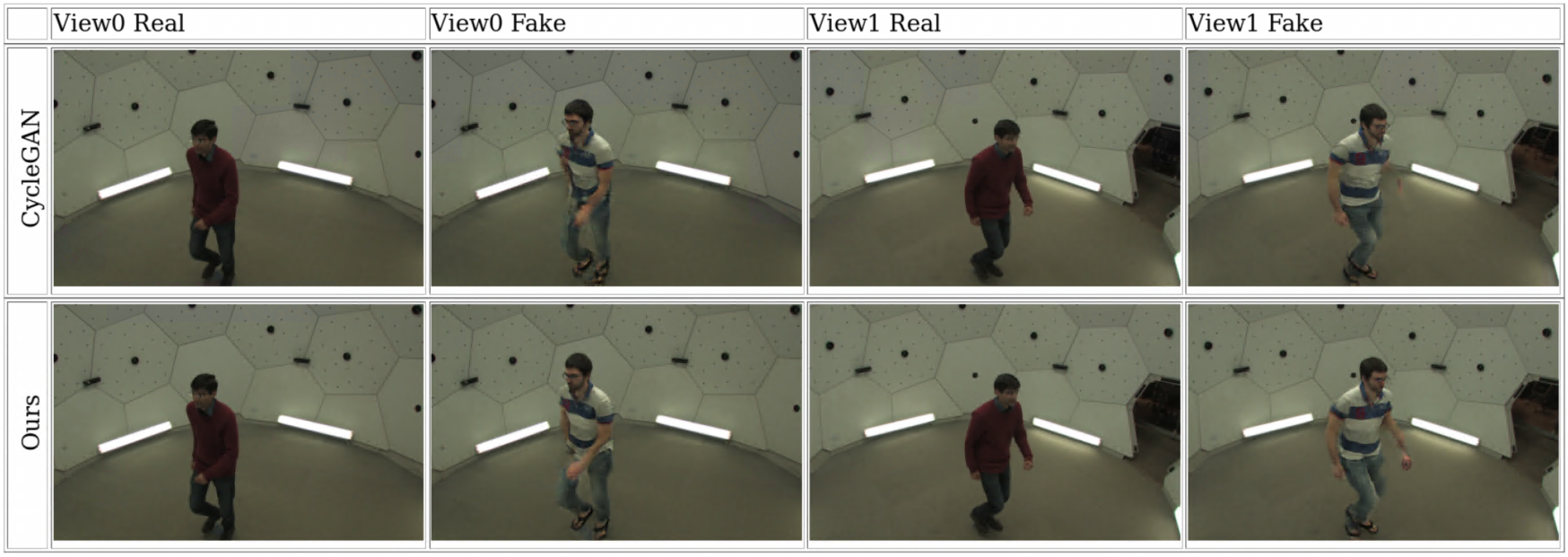}} \\
\imagecenter{\footnotesize c} &  \imagecenter{\includegraphics[trim={0 -15 0 40}, clip, width=0.9\textwidth]{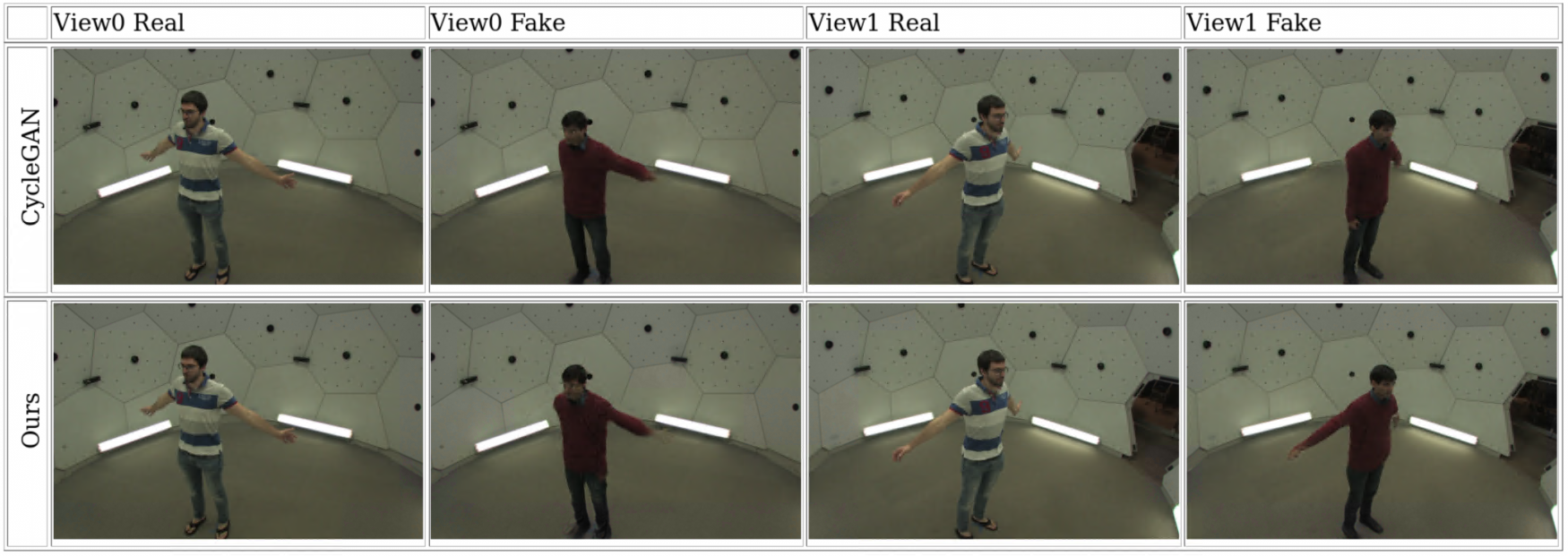}} \\
\imagecenter{\footnotesize d} &  \imagecenter{\includegraphics[trim={0 -15 0 40}, clip, width=0.9\textwidth]{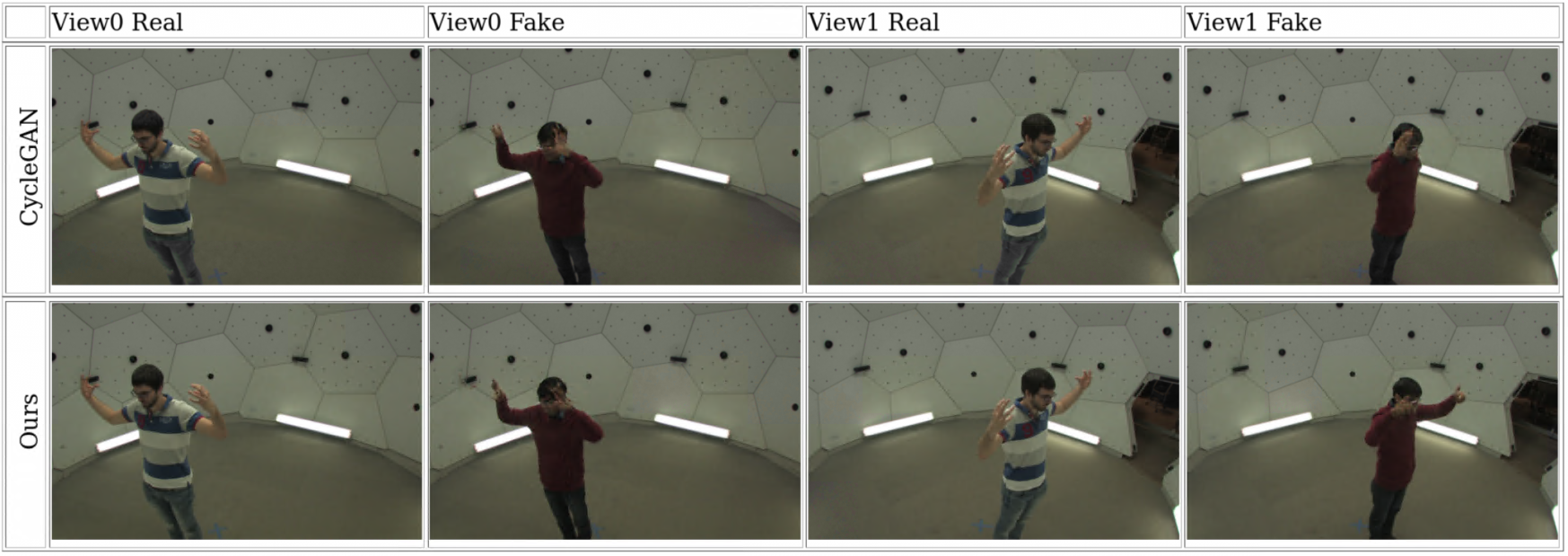}} \\
\end{tabular}
\caption{\textbf{Pose consistency in other camera viewpoints.} The examples demonstrate the effectiveness of our method for pose preservation in comparison to CycleGAN \protect\cite{CycleGAN2017}.
} 
\label{Fig:results_views_11_21_pair1_pose_preservation}
\end{figure}

\begin{figure}[t!]
\centering
 \begin{tabular}{ p{0.001\textwidth} p{0.9\textwidth} }
\imagecenter{\footnotesize a} &  \imagecenter{\includegraphics[trim={0 -15 0 0}, clip, width=0.9\textwidth]{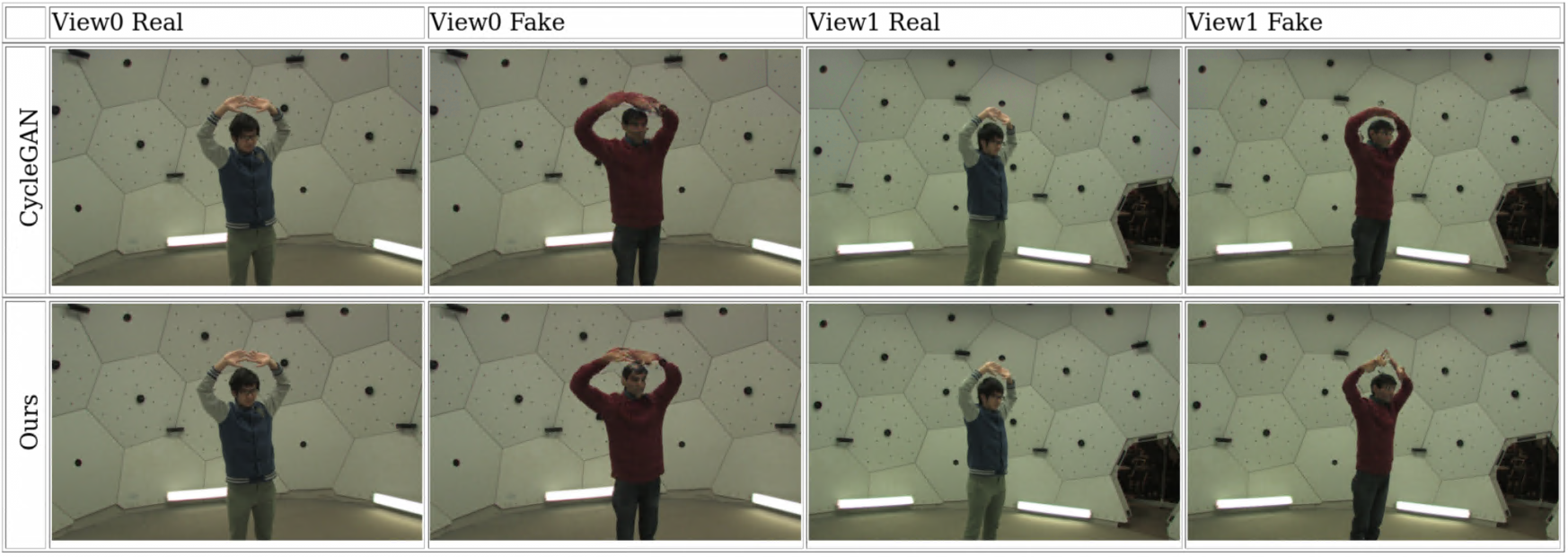}} \\
\imagecenter{\footnotesize b} &  \imagecenter{\includegraphics[trim={0 -15 0 40}, clip, width=0.9\textwidth]{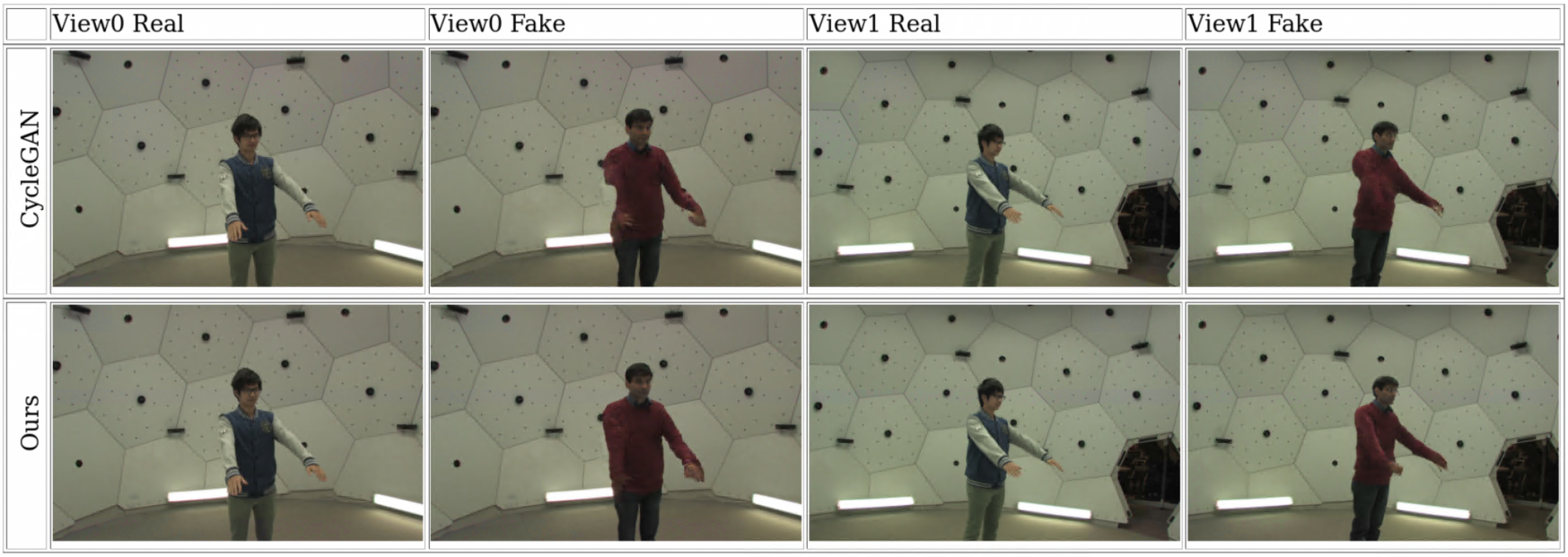}} \\
\end{tabular}
\caption{\textbf{Pose consistency in another pair of persons.}  The examples demonstrate the effectiveness of our method for pose preservation in comparison to CycleGAN \protect\cite{CycleGAN2017}.} 
\label{Fig:results_views_00_24_pair2_atob_pose_preservation}
\end{figure}

\vspace{0.1cm}
Visual results of our proposed method in comparison to CycleGAN are shown in Figures \ref{Fig:results_views_00_24_pair1_a2b_pose_preservation}-\ref{Fig:results_views_00_24_pair2_atob_pose_preservation}. Overall, the images generated by our model exhibit better translation quality of the human pose than the baseline due to the multi-view joint-learning imposed by the 3D constraints. 

\paragraph{The CMU-Panoptic Dataset.} CMU-Panoptic \cite{joo2016panoptic} is a multi-camera dataset that provides 30 Hz Full-HD video streams of 40 subjects (persons) from up to 31 synchronized cameras and a diverse set of annotations on the subjects. Among the available annotations is the $3D$-skeleton of each subject as acquired via triangulation using all camera views and is specified by the spatial positions of 19 body keypoints. In each of the tests described below, we use a pair of cameras and a pair of subjects along with a subset of 17 out of the 19 available 3D keypoint positions to coincide with the standard COCO format \cite{lin2015microsoft}.

\paragraph{Pose Consistency Across Views.}
Figures \ref{Fig:results_views_00_24_pair1_a2b_pose_preservation}-\ref{Fig:results_views_00_24_pair1_btoa_pose_preservation} demonstrate the effectiveness of our approach in generating images that exhibit better translation quality of the human pose than CycleGAN. Specifically, in Figure \ref{Fig:results_views_00_24_pair1_a2b_pose_preservation}a CycleGAN failed to preserve the pose of person $A$ in the translation to person $B$ for both camera viewpoints. Moreover, the 2D poses of person $B$ in the two fake images are totally inconsistent with one another. In contrast, our method managed to synthesize images in which the pose of person $B$ in view 0 is consistent with the pose of person $A$ in the real images and is spatially similar (but not identical) to the pose in view 1. In Figures  \ref{Fig:results_views_00_24_pair1_a2b_pose_preservation}b-c CycleGAN managed to preserve the pose of person $A$ in the fake images of view 0, but failed to do the same for view 1. In contrast, due to the joint learning imposed by the 3D constraints, our method better preserved the human pose for both views. Figure \ref{Fig:results_views_00_24_pair1_btoa_pose_preservation} shows results for the opposite translation from person $B$ to $A$.

\paragraph{Local Feature Preservation.} Figure \ref{Fig:results_views_00_24_pair1_a2b_feature_consistency} shows examples in which both CycleGAN and our approach managed to preserve the overall human pose for both camera viewpoints. However, local features, such as fingers, are more consistent between real images and fake images generated by our approach. We obtained these improved results despite the fact that the human pose consists of only 17 keypoints along the whole body \cite{lin2015microsoft} and none of them is on fingers.
These local feature improvements are likely to occur due to context completion of distorted body parts. The context completion enables better prediction of nearby keypoints, which reduces the 3D loss term. For example, visual details of fingers improve the recognition of hand keypoints.

\paragraph{Robustness and Stability.} Both Figures \ref{Fig:Sequence_comparison} and \ref{Fig:results_CycleGAN_collapse_cases} show the robustness and stability of our approach in comparison to CycleGAN \cite{CycleGAN2017}. Figure \ref{Fig:Sequence_comparison} demonstrates this behavior on sequences of video frames. Specifically, Figures \ref{Fig:Sequence_comparison}a-b show that the human pose in all fake images generated by CycleGAN for a sequence of real frames with different poses are incorrect. In contrast, using our model resulted with a better preservation of the pose compared to the ground-truth in a consistent manner for all frames in the sequence. Figure \ref{Fig:Sequence_comparison}c shows that for very similar real frames, CycleGAN synthesized images that are inconsistent in terms of human pose (different fake poses were generated for very similar real poses). In contrast, our model generated fake poses that better preserve the human pose in the real images in a consistent manner for all frames. The three examples show that our method produced temporally coherent appearance that is consistent with the source video frame, despite the fact that no temporal information was used and the fake frames were generated independently.

Figure \ref{Fig:results_CycleGAN_collapse_cases} shows an incorrect consistent behaviour of CycleGAN that is overcome by our approach. Specifically, the figure presents cases in which CycleGAN resulted in a similar mistake in its synthesis for different inputs, while our method synthesized the poses correctly. Figure \ref{Fig:results_CycleGAN_collapse_cases}a shows that for two different inputs CycleGAN synthesized the same incorrect composition of the hands in front of the body. In contrast, correct poses were obtained by our method for both cases. Figure \ref{Fig:results_CycleGAN_collapse_cases}b shows that in all three fake images synthesized by CycleGAN the right hand is mistakenly located on the side of the body as opposed to our results in which the hand is located correctly. In Figure \ref{Fig:results_CycleGAN_collapse_cases}c CycleGAN generated the same incorrect pose for two different inputs, while our model resulted with a correct pose even on the small details of the right hand fingers.

\paragraph{Additional Views and Subjects.} Figure \ref{Fig:results_views_11_21_pair1_pose_preservation} shows the effectiveness of our method compared to CycleGAN \cite{CycleGAN2017} in two other camera viewpoints. In Figures \ref{Fig:results_views_11_21_pair1_pose_preservation}a-b CycleGAN managed to preserve the pose on view 0, but failed on view 1. In contrast, the joint-learning of both views in our approach helped in generating fake images that are view-consistent and both preserve the human pose of the real images. Figures \ref{Fig:results_views_11_21_pair1_pose_preservation}c-d show similar results for the opposite person translation. Figure \ref{Fig:results_views_00_24_pair2_atob_pose_preservation} shows the performance of our method in comparison to CycleGAN \cite{CycleGAN2017} with a different pair of persons.

\section{Conclusion}

The proposed method tackles the problem of multi-view image-to-image translation of a foreground person that transfers the human appearance while preserving the pose across all views. 
We showed promising results achieved by the addition of 3D pose supervision into a CycleGAN framework to encourage pose-consistency across views. Additionally, the results demonstrate the robustness of the method and its potential in preserving local features near anatomical body keypoints. 

{\small
\bibliographystyle{abbrv}
\bibliography{references}
}

\end{document}